\pdfoutput=1

\documentclass[11pt]{article}

\usepackage[preprint]{acl}

\usepackage{times}
\usepackage{float}
\usepackage{amsmath}
\usepackage{latexsym}
\usepackage{xcolor}
\usepackage{hyperref}

\usepackage[T1]{fontenc}

\usepackage[utf8]{inputenc}

\usepackage{microtype}

\usepackage{inconsolata}
\usepackage{multirow}
\usepackage{graphicx}
\DeclareCaptionLabelFormat{andtable}{#1~#2  \&  \tablename~\thetable}

\title{Towards Fair and Efficient De-identification: Quantifying the Efficiency and Generalizability of De-identification Approaches}

\author{
Noopur Zambare\textsuperscript{1} \quad
Kiana Aghakasiri\textsuperscript{1} \quad
Carissa Lin\textsuperscript{1} \\[0.6em]
\textbf{Carrie Ye\textsuperscript{2,4}} \quad
\textbf{J. Ross Mitchell\textsuperscript{1,2,3}} \quad
\textbf{Mohamed Abdalla\textsuperscript{1,2,3}}
\\[0.6em]
\textsuperscript{1}Department of Computing Science, University of Alberta \\
\textsuperscript{2}Department of Medicine, University of Alberta\\
\textsuperscript{3}Alberta Machine Intelligence Institute (Amii)\\
\textsuperscript{4}Arthritis Research Canada\\[0.6em]
\texttt{{zambare, kaghakas, carissa1, cye, jmitche2, mabdall2}@ualberta.ca}
}

\begin{document}
\maketitle
\begin{abstract}
Large language models (LLMs) have shown strong performance on clinical de-identification, the task of identifying sensitive identifiers to protect privacy.  However, previous work has not examined their generalizability between formats, cultures, and genders. In this work, we systematically evaluate fine-tuned transformer models (BERT, ClinicalBERT, ModernBERT), small LLMs (Llama 1-8B, Qwen 1.5-7B), and large LLMs (Llama-70B, Qwen-72B) at de-identification. We show that smaller models achieve comparable performance while substantially reducing inference cost, making them more practical for deployment. Moreover, we demonstrate that smaller models can be fine-tuned with limited data to outperform larger models in de-identifying identifiers drawn from Mandarin, Hindi, Spanish, French, Bengali, and regional variations of English, in addition to gendered names. To improve robustness in multi-cultural contexts, we introduce and publicly release BERT-MultiCulture-DEID, a set of de-identification models based on BERT, ClinicalBERT, and ModernBERT, fine-tuned on MIMIC with identifiers from multiple language variants. Our findings provide the first comprehensive quantification of the efficiency-generalizability trade-off in de-identification and establish practical pathways for fair and efficient clinical de-identification.

Details on accessing the models are available at: \url{https://doi.org/10.5281/zenodo.18342291}
\end{abstract}

\section{Introduction}
De-identification is the process of removing personally identifiable information (PII) from data to protect individual privacy.
This step is crucial in healthcare research, where clinical texts often contain sensitive details such as patient names, addresses, contact information, and other identifiers. Regulations like the Health Insurance Portability and Accountability Act (HIPAA) \citep{hipaa1996} and the General Data Protection Regulation (GDPR) \citep{gdpr2016} require the de-identification of such clinical notes before sharing. However, protecting this sensitive information is challenging, leading healthcare organizations to restrict sharing such data.

\begin{figure}[t]
\includegraphics[width=\columnwidth]{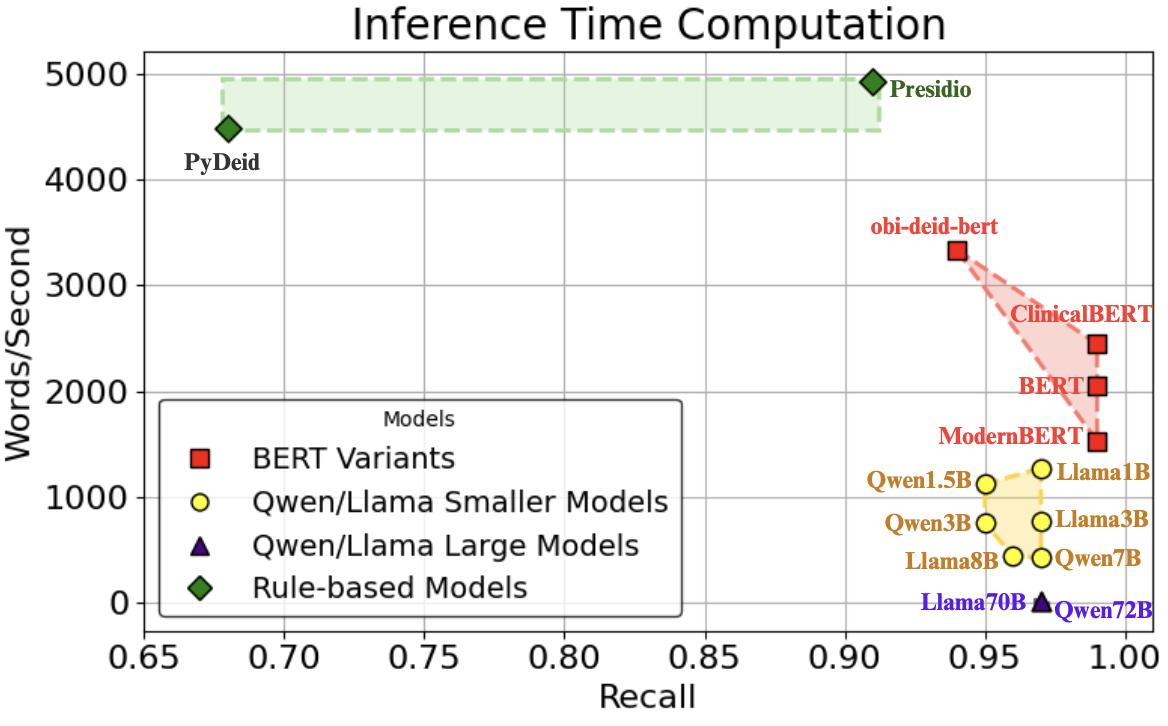}
\caption{Inference time computation of different de-identification models. The y-axis shows words processed per second, and the x-axis shows recall. Results are grouped by model type: BERT variants, smaller Qwen(1.5-7B)/Llama(1-8B) models, large Qwen-72B/Llama-70B models, and rule-based models.}
\label{fig:inference}
\end{figure}

Traditionally, de-identification was performed using dictionary or rule-based systems, which rely on predefined patterns or lists of dictionaries to identify PII. Although these methods are computationally inexpensive, they often fail to generalize between note formats or variations in identifier style, particularly when dealing with identifiers from different cultural contexts or  institutions \citep{ferrandez2012generalizability}.

To improve the generalizability of de-identification, researchers have adopted transformer-based models such as BERT \citep{devlin2019bert} and ClinicalBERT \citep{alsentzer2019publicly}. These models have demonstrated substantial improvements in de-identification performance. Using contextual embeddings, these models can detect sensitive tokens with high precision and recall, even in complex narrative structures. However, models like BERT require fine-tuning for each type of identifier (e.g., names, addresses, phone numbers), and this process requires labeled data.

To minimize the need for training data (generated by the data owners looking to de-identify their data), researchers have more recently experimented with using LLMs such as Llama \citep{grattafiori2024llama}, Qwen \citep{bai2023qwen, team2024qwen2}, Mixtral \citep{jiang2024mixtral}, and ChatGPT \citep{radford2018improving}. Recent work suggests that these models perform well if not better than BERT-based models \citep{altalla2025evaluating, wiest2025deidentifying, pissarra2024unlocking}.

Although LLM-based de-identification has demonstrated high performance, certain critical challenges limit its adoption by healthcare institutions. First, most academic research on LLM-based de-identification experiment with very large models, which are difficult to deploy locally for many institutions. Furthermore, even if institutions had the computational ability to run these models, inference with large models remains computationally intensive and slow, making them impractical for clinical deployment (though this has not been formally quantified in any previous work). Given the constraints of real-world clinical settings, understanding this trade-off is crucial.

At the same time, while some studies have examined cross-format or cross-lingual performance, comprehensive evaluations of LLM-based de-identification models across culturally diverse identifiers, different formats, and gendered identifiers are lacking, limiting understanding of their robustness and real-world applicability.

Taken together, these gaps raise two critical questions: 1) How (in-)efficient is LLM-based de-identification and can smaller LLM de-identification models achieve performance comparable to very LLMs (e.g. 70B parameters) while substantially reducing computational requirements and inference time?, and 2) Are LLM-based de-identification models able to generalize across healthcare institutions, genders and language/cultural variations in identifiers?

To address these questions, our work makes the following contributions:
\begin{itemize}

    \item \textbf{Efficient De-identification:} We conduct a systematic analysis of inference efficiency and performance across fine-tuned BERT models (BERT, ClinicalBERT, ModernBERT) as well as smaller Llama (1B, 3B, 8B) and Qwen (1.5B, 3B, 7B) models. We demonstrate performance comparable to large models (Llama 3.3-70B and Qwen2.5-72B) with substantially lower compute requirements and faster inference time.  
    
    \item \textbf{Generalizability:} 
    We evaluate model robustness across formats, gendered identifiers, and culturally diverse identifiers drawn from Mandarin, Hindi, Spanish, French, and English naming traditions. Our goal is to test whether English models remain reliable when the identifiers reflect the cultural diversity seen in large urban anglophone cities. Our results highlight the limitations of all models in adapting to diverse real-world contexts.

    \item \textbf{Multi-Cultural Deid:} To address substantial performance degradation in multicultural identifiers, we fine-tuned BERT, ClinicalBERT, and ModernBERT, developing BERT-MultiCulture-DEID, a set of models that exhibit improved multi-cultural generalization.

\end{itemize}

\section{Related Work}
De-identification has been an active field of research for over 30 years \cite{sweeney1998strategies}. During this time, the field has continuously adopted the most advanced methods, from regular expressions \cite{neamatullah2008automated} to traditional machine learning algorithms \citep{liu2017identification}, and transformer-based models (e.g., BERT and ClinicalBERT) \cite{johnson2020deidentification}. Consequently, the performance of automated de-identification systems has continued to improve with performance metrics (e.g., precision and recall) nearing perfection \citep{johnson2020deidentification, moore2023transformer}. Most recently, LLMs (e.g., Llama, Qwen, Mixtral) have been adopted for de-identification \citep{pissarra2024unlocking, altalla2025evaluating, wiest2025deidentifying} and have demonstrated near-perfect recall without requiring (much) labeled training data, demonstrating impressive zero-shot and few-shot generalization \citep{brown2020language}.

\subsection{Efficiency in Clinical De-identification}
De-identification is a time-consuming task for humans. \citet{dorr2006assessing} measured that de-identifying a clinical note required 87.3 $\pm$ 61 seconds, while \citet{douglass2004computer} found that humans were able to de-identify between 250 and 350 words per minute.

\citet{heider2020comparative} conducted a comparative evaluation of de-identification systems (Amazon Comprehend Medical PHId \citep{aws_comprehend_medical}, Clinacuity \citep{meystre2023clinideid}, and the National Library of Medicine’s (NLM) Scrubber \citep{nlm_scrubber} and concluded that none of the systems simultaneously achieved optimal performance in both speed and accuracy. NLM Scrubber achieved a recall of 0.47 and processed 4.09 notes/sec, CliniDeID achieved a recall of 0.99 at 0.29 notes/sec, and Amazon Comprehend Medical PHId achieved a recall of 0.80 at 1.75 notes/sec.

As de-identification models have grown more intricate, their inference time has also increased. To address increases in compute time, \citet{sundrelingam2025pydeid} introduced pyDeid for rapid, generalizable rule-based de-identification, showing competitive speed with an average runtime of 0.48 seconds/note with a best recall of 0.95, outperforming traditional tools such as Deid (0.93 seconds/note, recall: 0.87; \cite{neamatullah2008automated}) and Philter (6.38 seconds/note, recall: 0.92; \cite{norgeot2020protected}). Other researchers have sought to enhance the efficiency of LLMs by fine-tuning smaller models. \citet{doremus2025harnessing} demonstrated that small LLMs (e.g., 7B) can be fine-tuned to achieve high de-identification performance (F1 score: 0.97 and recall: 0.93), although they did not specifically measure inference time. 
\citet{chen2025benchmarking} and \citet{naguib2024few} evaluated pretrained models on biomedical NLP tasks (performance and cost) and NER (performance and carbon emissions), respectively.

Current literature exhibits several shortcomings. First, existing efficiency assessments use different datasets, which hinders direct comparability. Second, these evaluations lack uniformity in metrics; measuring time per note complicates future comparisons due to variability in note length. Finally, there have been no direct comparisons of the efficiency of LLMs with other competing models.

\subsection{Generalization and Robustness}
Recent advances in pre-trained transformer-based models, such as variants of BERT and GPT, Llama, etc., \citep{radford2018improving} have demonstrated strong generalization capabilities in various NLP tasks \citep{budnikov2025generalization}. BERT-based models typically excel in domain-specific fine-tuning due to their bidirectional contextual representations, whereas GPT-style models benefit from autoregressive pre-training that allows effective few-shot and zero-shot generalization \citep{brown2020language}.

Recognizing potential concerns about robustness, researchers have explored the generalizability of clinical de-identification models. \citet{xiao2023name} identified significant performance disparities in most de-identification approaches they evaluated, with biases evident in demographic dimensions such as gender and race. \citet{chen2024examining} analyzed pre-trained transformer models in discharge summaries and nursing notes, discovering significant accuracy declines due to differences in structure and PII entity distributions. \citet{kim2024generalizing} observed that state-of-the-art de-identification models show poor generalization on new datasets, mainly due to challenges in maintaining training corpora, as well as variations in labeling standards and patient record formats across institutions. They also suggested that GPT models (and large LLMs more generally) could help address these challenges, an assumption which our analysis shows is incorrect. 

Unfortunately, current research has several limitations. First, assessments are frequently confined to smaller models, as illustrated by \citet{chen2024examining}, who restricted their evaluation to BERT-based models. Second, previous studies often limit their evaluations to particular types of PII. For example, \citet{xiao2023name} focused solely on the robustness of the models with respect to patient names. However, in practice, other identifiers, such as addresses (including postal codes and street patterns), institution or hospital names, phone numbers, and other entities, also vary across different contexts.

Finally, while these studies perform fine-tuning of transformer-based models for de-identification, they do not analyze how fine-tuning impacts robustness in smaller models across formats, genders, and identifiers drawn from different language variants.

\section{Datasets}
We performed experiments using two English-language clinical datasets: the publicly available MIMIC-III dataset \citep{johnson2016mimic} and a smaller proprietary dataset consisting of rheumatology referral letters. In this work, we define PII as any token belonging to one of the following categories: name, phone/fax, hospital, city, state, address, country, company, university, date, email, and other (MRN, account number, etc.). Appendix Table~\ref{tab:data_stats} presents some descriptive statistics of the two datasets, and Figure~\ref{fig:pii_distribution} shows the statistics of identifiers in the test data.

All datasets, codes and models utilized in this study were collected and used in compliance with their respective licenses and access requirements.

\subsection{MIMIC-III}
We targeted PII identified in the Health Insurance Portability and Accountability Act (HIPAA) for MIMIC-III. These include patient names, telephone numbers, addresses, and dates. We sampled 4,000 discharge summaries from the MIMIC-III dataset. Of these, 2,000 were used for fine-tuning with varying subset sizes (250, 500, 1,000, and 2,000 notes), 1,000 were used for validation, and 1,000 were sequestered for testing.

\subsection{Private Clinical Dataset (PCD) - Alberta Health Services}
This dataset consists of 204 referral letters from physicians to rheumatologists, covering a wide range of clinical scenarios. All notes were saved in PDF format and consisted of a combination of digitally typed documents, scanned files, and handwritten letters. We extracted text from these PDFs using optical character recognition (OCR) with the Doctr library \citep{doctr2021}. Authors then manually removed all sensitive information, including patient names, provider names, phone numbers, addresses, medical record numbers, account numbers, other identifiers, dates, and any additional personally identifiable information. The collection and use of this dataset was approved by the University of Alberta's REB (\#Pro00141020).

\subsection{Faker}
We substituted the de-identified masks in the original notes with realistic surrogate data, preserving the structure and readability of the clinical text while ensuring that no real patient information was exposed. Replacement was performed using the Python-based Faker library \citep{Faraglia_Faker}. This library provides a wide variety of synthetic identifiers, including names, addresses, phone numbers, email addresses, organization names, dates, and other identifiers such as medical record numbers and account numbers. For all experiments, except where specified, we used the default settings (e.g., using the US-locale).

\section{Models}
We tested multiple models: pretrained models including Llama 3.1-8b, Llama 3.2-1B, Llama 3.2-3B, Llama 3.3-70B, and Qwen 2.5 (1.5B, 3B, 7B, and 72B), BERT, ClinicalBERT, ModernBERT, and three open source toolkits (obi-deid-bert \citep{obi_deid_bert_i2b2}, pyDeid \citep{sundrelingam2025pydeid}, and Presidio \citep{presidio}).

\subsection{BERT, ClinicalBERT and ModernBERT}

BERT is a transformer-based language model pre-trained on large corpora using a masked language modeling objective, enabling it to capture rich contextual representations of text \cite{devlin2019bert}. ClinicalBERT is a variant of BERT that is further pre-trained on clinical notes and discharge summaries from the MIMIC III dataset \cite{huang2019clinicalbert}. This improves ClinicalBERT's ability to identify medical terminology and context. ModernBERT \citep{warner2024smarter} is a more recent adaptation of BERT, offering better performance, efficiency, and longer sequence handling compared to standard BERT models.

In the experiments below, we fine-tuned various BERT, ClinicalBERT, and ModernBERT models as comparators. In this section, we describe generalized fine-tuning information used for all varieties. The specifics for each variation are described in their respective sections. The models were fine-tuned to perform binary token classification or multiclass token classification. For binary classification, each token was classified as PII or non-PII. For multiclass classification, each token was classified as non-PII or as a member of a set with multiple PII types. We focus on the results of binary PII prediction, with multiclass results presented in Appendix \ref{sec:multi-class}.

To ensure a fair comparison across different dataset sizes (250, 500, 1000, and 2000 samples), the number of training epochs was kept the same for all models. We selected 10 epochs because of stabilization in validation loss at this number.

We also included \emph{obi-deid-bert}\citep{obi_deid_bert_i2b2}, a publicly available de-identification model based on ClinicalBERT (pre-trained on MIMIC-III and fine-tuned on i2b2/n2c2 \citep{stubbs2015annotating}), as a baseline.

\subsection{LLMs (Llama and Qwen)}
LLMs are transformer-based models pretrained using the next-token prediction objective, giving rise to generative capabilities. Llama \cite{grattafiori2024llama} and Qwen \cite{bai2023qwen} are families of open-weight language models.

We fine-tuned the Llama (1B, 3B, 8B) and Qwen (1.5B, 3B, 7B) models for token classification. In this work, we used them exclusively for token-level classification. A classification head was appended to the final transformer layer, and fine-tuning was restricted to Low-Rank Adaptation (LoRA) and the classification head. Small models (1B/1.5B) were fine-tuned for 5 epochs, medium-sized models (3B) for 5 epochs, and large models (7B and 8B) for 3 epochs. Detailed experimental setup is explained in the Appendix \ref{sec:fine-tuning-setup}.

\subsection{Prompt-tuning}
We prompt-tuned the largest variants, Llama-70B and Qwen-72B, using one-shot prompting on a small subset of five MIMIC-III notes, using the prompt mentioned in Appendix Figure \ref{fig:prompt_mimic}, detalied in Appendix Section \ref{sec:prompts}.

\subsection{Presidio}
Microsoft Presidio \citep{presidio}, an open source framework, detects PII using named entity recognition (NER) techniques.

\subsection{PyDeid}
We evaluated PyDeid \citep{sundrelingam2025pydeid}, a rule-based system that uses regular expressions and fixed inclusion/exclusion lists to de-identify free-text clinical data.

\section{Metrics}
\subsection{Standard Classification Metrics}
For the majority of our analyses, we present precision and recall -- standard classification metrics used in de-identification. Precision is the proportion of tokens predicted as PII that are actually PII. Recall is the proportion of PII in a note that is correctly flagged as PII.

\subsection{Clinical Information Retention}
Recent work \cite{aghakasiri2025not} has demonstrated that standard classification metrics do not provide a complete picture of model performance. Specifically, when a false positive occurs, clinically relevant information may be removed. This reduces the utility of notes and thus negates the point of de-identification. Balancing the preservation of clinical utility with data sensitivity is critical. Therefore, in this work, we use the CIRE metric proposed by \citet{aghakasiri2025not}, which uses an LLM to measure retention of clinical information by calculating the proportion of sentences that have changed clinically relevant information (CIRE prompt is presented in Appendix Figure~\ref{fig:CIRE}).

\subsection{Inference Efficiency}
We also report the number of words/second. We prioritize this metric over the commonly references seconds/note, as the latter depends on the note length, which varies.

\section{Experiments and Results}
\begin{figure*}[t]
\centering
\includegraphics[width=0.8\textwidth]{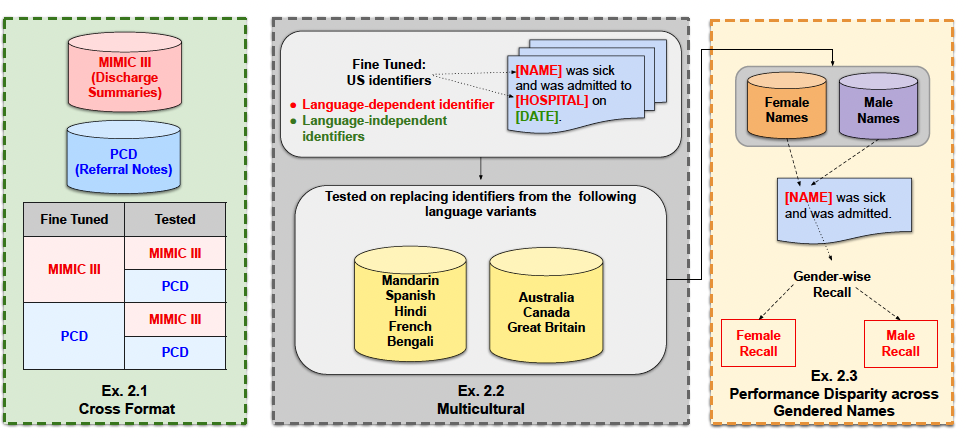}
\caption{Setup of generalization testing. \textbf{(Ex. 2.1) Cross-format testing:} Models were fine-tuned on MIMIC and tested on both MIMIC and PCD, and vice versa, fine-tuned on PCD and tested on both datasets. \textbf{(Ex. 2.2) Multi-cultural testing:} Models fine-tuned on MIMIC notes with US English identifiers were tested on notes with identifiers from different language variants. \textbf{(Ex. 2.3) Performance disparity across gendered names:} Notes with identifiers from different language variants were used, and recall was evaluated specifically for name identifiers.}
\label{fig:ex2}
\end{figure*}

\begin{table}[t]
\tiny
\centering
\begin{tabular}{lllllllll}
\hline
\textbf{Model}  & \textbf{P} & \textbf{R}  & \textbf{CIRE} & \textbf{Time (s)} & \textbf{STD
 (s)} & \textbf{Words/sec}\\
\hline

BERT               & 0.99 & 0.99    & 0.99& 67 & 2.0 & 2048   \\
ModernBERT          & 0.99 & 0.99  & 0.99& 90 & 3.1 & 1528   \\
ClinicalBERT         & 0.99 & 0.99  & 0.99& 56 & 1.8 & 2446 \\
\hline
Qwen-1.5B & 0.96 & 0.94 & 0.99& 122 & 2.5 & 1121\\
Qwen-3B & 0.96 & 0.95 & 0.99& 182 & 2.8 & 751\\
Qwen-7B & 0.96 & 0.95 &0.99 & 308 & 4.4 &443 \\
Llama-1b      &  0.97 & 0.96   & 0.99&108 & 3 & 1271 \\
Llama-3b      & 0.97 & 0.97 & 0.99& 176 & 5.4 & 776\\
Llama-8b      & 0.97 & 0.97    &0.99 & 322 & 6.2 & 423\\
Llama-70B* & 0.75   & 0.97  &0.98 & 12870& - & 10  \\
Qwen-72B* & 0.80 & 0.97 & 0.98 &16315  &- & 8\\
\hline
pyDeid &  0.67 & 0.68 & 0.99& 31 & 0.2 & 4490\\
obi-deid-bert & 0.91 & 0.94 &0.99 & 41 & 1.4  & 3333\\
Presidio & 0.61 & 0.91 & 0.89&  28 & 0.1 & 4931\\

\hline
\end{tabular}
\caption{\textbf{Experiment 1:} Performance and GPU-based inference time for various de-identification models. We report the words-per-second rate, along with the time required to process 100 notes and the standard deviation across 10 runs to de-identify 100 notes. *Evaluated only once due to compute cost. \textbf{P: Precision, R: Recall}}
\label{tab:inference_time_models}
\end{table}

\subsection{Experiment 1: LLM-based de-identification}

Inference performance was measured on an NVIDIA A100-SXM4-80GB GPU. All models (except Llama-70B and Qwen-72B) were fine-tuned on 1,000 MIMIC-III discharge summaries. Llama-70B and Qwen-72B were prompt-tuned rather than fine-tuned on five discharge notes. For all fine-tuned models, inference time was computed on a test set of 100 discharge notes, repeated 10 times.
The inference compute time of prompt-tuned models was measured only once due to the high inference time. 

Table~\ref{tab:inference_time_models} highlights clear trade-offs between model size, performance, and computational efficiency. BERT, ModernBERT and ClinicalBERT achieved near-perfect precision and recall (0.99 each) with relatively low inference times ranging from 1528 to 2446 words/sec. For LLMs, the smaller variants (1B–8B) achieved high precision and recall (0.96–0.97) with inference ranging from 423 to 1271 words/sec. In contrast, Llama-70B and Qwen-72B processed 100 notes in more than 12,000 seconds (more than 3 hours), achieving only 8 to 10 words/sec, even when using two GPUs. Although they achieved high recall, their precision was not competitive, underscoring the limited practicality of large LLMs for de-identification. For subsequent experiments, we dropped Qwen2.5-72B, as its performance was comparable to Llama3.3-70B but was much slower during inference.

Rule-based systems exhibited the fastest runtimes, but at the expense of performance. Presidio achieved the fastest runtime (4931 words/sec) but had poor precision (0.61). Similarly, pyDeid was efficient (4490 words/sec), but underperformed in precision and recall (0.67 and 0.68). In contrast, obi-deid-bert (fine-tuned on i2b2) offered a more balanced trade-off, with moderate inference time (3333 words/sec) and higher recall (0.94). All models performed well in the CIRE score with no meaningful differences between model classes (except Presidio, which was about ~0.10 lower than all other models).

\subsection{Experiment 2: Generalization}

We conducted three experiments (cross-format, multi-cultural and performance disparity across gendered names) to evaluate the robustness and generalization of de-identification models, illustrated in Figure \ref{fig:ex2}.

\begin{table}[t]
\tiny
\centering
\begin{tabular}{lcc|cc|cc|cc}
\hline
\multirow{1}{*}{\textbf{Train - Test}} & 
\multicolumn{2}{c|}{\textbf{PCD-PCD}} & 
\multicolumn{2}{c|}{\textbf{PCD–M}} & 
\multicolumn{2}{c|}{\textbf{M–M}} & 
\multicolumn{2}{c}{\textbf{M–PCD}} \\
 \textbf{Model}& P & R & P & R & P & R & P & R \\
\hline
BERT         &  0.99 & 0.99  &  0.99 & 0.98  &    0.99 & 0.99   & 0.99  &  0.98 \\
Qwen-7b      &  0.95 & 0.90  & 0.71  &  0.64 &   0.96 & 0.95    &  0.93 &  0.90 \\
Llama-8b     &  0.98 &  0.97 &  0.93 & 0.85  &   0.98 & 0.97   & 0.96  &  0.94 \\
Llama-70b*   & 0.92  & 0.98  &  0.70 & 0.97  & 0.72  & 0.97  & 0.93  &   0.97\\
\hline
\end{tabular}
\caption{\textbf{Experiment 2.1 - Cross-format Evaluation:} Models fine-tuned on Private Clinical Dataset and MIMIC-III separately and tested on both datasets. *Prompt-tuned. \textbf{P: Precision, R: Recall, M: MIMIC} }
\label{tab:cross_institute_test}
\end{table}

\subsubsection*{Experiment 2.1: Cross-Format Generalization}

\begin{figure*}[t]
  \begin{minipage}{\textwidth}
  \begin{minipage}[b]{0.49\textwidth}
    \scriptsize
    \centering
    \begin{tabular}{lc|c|c|c|c}
    \hline
    \textbf{Model} 
    & \textbf{Mandarin}
    & \textbf{Spanish}
    & \textbf{Hindi}
    & \textbf{French}
    & \textbf{Bengali} \\
    \hline
    BERT               & 0.80  & 0.77  & 0.80  & 0.83  & 0.75 \\
    ModernBERT      & 0.75  &0.83  & 0.92  & 0.90  & 0.81\\
    ClinicalBERT       & 0.87  & 0.84  & \textbf{0.97}  & 0.87  & \textbf{0.97}\\
    \hline
    Qwen-1.5B  & 0.80  & 0.86  & 0.91  & 0.89  & 0.80\\
    Qwen-3B  & 0.65  & 0.82  & 0.90  & 0.89  & 0.89\\
    Qwen-7B  & 0.71  & 0.85  & 0.89  & 0.89  & 0.76\\
    Llama-1b          & 0.83 & 0.80 & 0.90  & 0.87  & 0.77 \\
    Llama-3b     & 0.71  & 0.82 & 0.88 & 0.87  & 0.76  \\
    Llama-8b     & 0.76 & 0.85 & 0.91 & 0.90 & 0.78     \\
    Llama70 &  \textbf{0.96} &  \textbf{0.95} &  \textbf{0.97} &  \textbf{0.97} & 0.96  \\
    \hline
    pyDeid & 0.58 & 0.64 & 0.62 & 0.66 & 0.59\\
obi-deid-bert & 0.94 &0.84 & 0.92 & 0.90 & 0.93\\
Presidio & 0.94 & 0.90  & 0.87 & 0.92 & 0.85\\
    \hline
  \end{tabular}
  \caption{\textbf{Experiment 2.2.1:} Recall of de-identification models fine-tuned on 1000 MIMIC-III and tested on 500 samples from five languages. \textbf{P: Precision, R: Recall}. Full results in Appendix \ref{app:full_results}.}
  \label{tab:cross_lingual_test}
    \end{minipage}
  \hfill
  \begin{minipage}[b]{0.49\textwidth}
    \centering
    \includegraphics[width=0.95\textwidth]{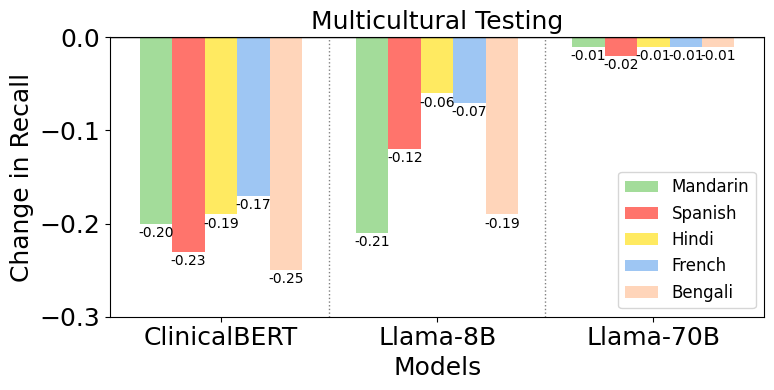}
    \caption{Relative difference in recall of the same model tested on US identifiers versus other languages. The models is fine-tuned on 1,000 samples with US English identifiers and evaluated on 500 samples for each language.}
    \label{fig:performance_across_languages}
  \end{minipage}
  \end{minipage}
\end{figure*}

To evaluate cross-format generalization, we fine-tuned the best performing models from Table~\ref{tab:inference_time_models} on 1,000 MIMIC-III discharge summaries and tested them on a separate collection of 1,000 MIMIC-III discharge notes (the MIMIC-test set) and 20\% of the private data set of referral notes. In contrast, we also fine-tuned the models using 80\% of the private dataset then tested them on the MIMIC-test set and 20\% of the private dataset. This setup allowed us to assess robustness across different types of clinical notes and narrative styles. The results of this evaluation are shown in Table~\ref{tab:cross_institute_test}.

BERT achieved consistently high performance across all train–test combinations, with precision and recall near 0.99, indicating strong generalization across formats. Llama-8B also demonstrated robust performance, though slightly lower in cross-format testing (PCD-M). This decline appears to be driven by variations in the narrative style between institutions. Surprisingly, Qwen-7B experienced a substantial drop in performance, nearly a 30-point drop evaluating on (PCD-M). Llama-70B, which was prompt tuned (with prompts for MIMIC-III and PCD provided in Figure~\ref{fig:prompt_mimic} and Figure~\ref{fig:prompt_pcd}, respectively), maintained high recall across datasets, but tended to over-mask, resulting in reduced precision in some cross-dataset scenarios.

\subsubsection*{Experiment 2.2: Multi-Cultural Generalization}

To evaluate how de-identification models trained on US English identifiers perform on identifiers from other language variants written in English (henceforth: multi-cultural generalization), instead of using the standard Faker settings, we replaced the identifiers using specialized settings in the Faker library. We picked the languages most spoken in the world (Mandarin, Spanish, Hindi, French, and Bengali)\footnote{Arabic, although among the top five languages, was excluded from the full evaluation due to limitations in the Faker library, which does not provide well-defined identifiers for all PII categories in Arabic.} \citep{ethnologue2025}. Note: unlike previous work, our changes were more comprehensive and included changes beyond just names to other identifiers (e.g., addresses, phone numbers, and other identifier types). 

This set of experiments used the MIMIC dataset. The models were fine-tuned on clinical notes with US English identifiers and then tested on notes with identifiers from other language variants.

Table~\ref{tab:cross_lingual_test} reports the recall of de-identification models trained in US-English and tested on identifiers from five non-English language variants (Mandarin, Spanish, Hindi, French, Bengali), with full results presented in Appendix Table \ref{tab:cross_lingual_test_appendix}. We found substantial recall drops for all models except for Llama-70B. More specifically, BERT, ModernBERT and ClinicalBERT maintained high precision, and ClinicalBERT achieved the highest recall for several language variants (Hindi and Bengali). Llama-1B and Llama-8B achieved strong recall, while prompt-tuned Llama-70B maintained very high recall but exhibited overmasking, leading to reduced precision.

Rule-based systems such as PyDeid and Presidio do not mask certain identifiers (e.g., hospital names, addresses, phone numbers) and were therefore evaluated differently. Only the identifiers these models actually mask were considered during the evaluation. For specific cases, the evaluation can be nondeterministic; for example, if a hospital name includes a personal name (e.g., `Jack's Clinic'), the model may mask `Jack' but not the full entity, which can lead to variations in reported metrics.

\begin{table*}[t]
  \tiny
  \centering
  \begin{tabular}{p{2.1cm}p{0.35cm}p{0.35cm}|p{0.35cm}p{0.35cm}|p{0.35cm}p{0.35cm}|p{0.35cm}p{0.35cm}|p{0.35cm}p{0.35cm}|p{0.35cm}p{0.35cm}|p{0.35cm}p{0.35cm}|p{0.35cm}p{0.35cm}}
    \hline
    \textbf{Model} 
    & \multicolumn{2}{c}{\textbf{Mandarin}} 
    & \multicolumn{2}{c}{\textbf{Spanish}} 
    & \multicolumn{2}{c}{\textbf{Hindi}} 
    & \multicolumn{2}{c}{\textbf{French}}
    & \multicolumn{2}{c}{\textbf{Bengali}}
    & \multicolumn{2}{c}{\textbf{GB}}
    & \multicolumn{2}{c}{\textbf{AU}}
    & \multicolumn{2}{c}{\textbf{CA}} \\
    \hline
    & P & R 
    & P & R 
    & P & R 
    & P & R 
    & P & R
    & P & R
    & P & R
    & P & R \\
    \hline
    BERT (all)         & 0.986 & 0.999  & 0.982 & 0.998 &  0.980 & 0.998 & 0.980 & 0.998 & 0.980 & 0.998 & 0.979 & 0.996 & 0.977 & 0.995 & 0.977 & 0.996 \\
    BERT (all–cult)    & 0.988 & 0.993 & 0.984 & 0.981 & 0.982 & 0.997 & 0.980 & 0.993 &0.981  & 0.995 & 0.982 & 0.975 & 0.982 & 0.988 & 0.982 & 0.990 \\
    ModernBERT (all)         & 0.984 & 0.998 & 0.978 & 0.997   &0.976  & 0.997 & 0.973 & 0.995 & 0.977 & 0.997 & 0.973 &  0.995 &0.971 & 0.993& 0.971 &0.994  \\
    ModernBERT (all–cult)    & 0.985 & 0.996 & 0.978 & 0.984 & 0.974 & 0.998 & 0.979 & 0.991 & 0.975 & 0.976 & 0.971 & 0.976 & 0.977 & 0.992 & 0.975 & 0.989 \\
    ClinicalBERT (all)       & 0.980 & 0.998 & 0.975 & 0.999 & 0.973 & 0.998 & 0.973 & 0.998 & 0.974 & 0.997 &  0.971& 0.995 & 0.970 & 0.994 &0.970  & 0.995 \\
    ClinicalBERT (all–cult)  & 0.982 & 0.951 & 0.980 & 0.993 &0.977  & 0.997 & 0.976 & 0.992 & 0.976 & 0.995 & 0.974 & 0.992 & 0.974 & 0.992 & 0.973 & 0.991 \\
    \hline
  \end{tabular}
  \caption{\textbf{Experiment 3:} Performance of different de-identification models fine-tuned on MIMIC-III across language variants. “all” = trained on all variants; “all–cult” = trained on all variants except the target variant. \textbf{P: Precision, R: Recall}}
  \label{tab:multi-deid}
\end{table*}

Appendix Table~\ref{tab:cross_dialect_test} presents the performance of models on identifiers from English variants (Great Britain, Australia, Canada). All fine-tuned transformer models maintained high precision and recall, demonstrating robustness to variations in addresses, phone numbers, and other identifiers in English-speaking regions. Prompt-tuned Llama-70B again showed high recall but decreased precision due to over-masking. These results indicate that, while models generalize well across English variations of identifiers, multi-lingual identifier variations remain more challenging, requiring careful model selection and tuning. Figure~\ref{fig:performance_across_languages} presents the results for the best performing models for this set of experiments.

\subsubsection*{Experiment 2.3: Performance disparity across Gendered Names}
We evaluated the de-identification models, fine-tuned on MIMIC-III with US-based identifiers, using names drawn from the same set of languages as in Experiment 2.2. To assess potential gender bias, we separately measured performance on masculine and feminine names, allowing us to analyze model robustness across gendered name variations. Specifically, we computed the recall for each gender.

Appendix Table~\ref{tab:cross_gender_test} summarizes the recall performance of de-identification models for feminine ($R_f$) and masculine ($R_m$) names across multiple languages (Mandarin, Spanish, Hindi, French, Bengali) and English variants (Great Britain). Almost all models achieved similar recall between genders, with differences generally below 0.05. Qwen variants exhibited a larger gap for French, where the difference in recall exceeded 0.05.

\subsection{Experiment 3: Developing BERT-MultiCulture-DEID}

As highlighted in Table \ref{tab:cross_lingual_test}, the obi-deid-bert model experienced substantial drops in model recall depending on the model evaluated (with up to a 10\% drop in recall in Spanish). Unfortunately, this is the only readily accessible publicly available de-identification model accessible to most data curators. To address this gap, we sought to explore the feasibility of improving the performance of BERT-based de-identification on identifiers from unseen language variants. 

For this experiment, we selected three BERT variants: BERT, ClinicalBERT, and ModernBERT (which can reasonably serve as replacements to the existing model). To perform this experiment, we fine-tuned two variations of each model. First, we trained models on all of the language variants present in Tables \ref{tab:cross_lingual_test} and \ref{tab:cross_dialect_test}. Second, we trained models on all variants except the one being evaluated (e.g., the Mandarin evaluation would be trained on all other language variants except for Mandarin). The observed difference in performance serves as an indicator of the generalization gap exhibited by the model.

Table \ref{tab:multi-deid} presents the results of this experiment. First, we observe that training on all language variants improves performance on all variants (since these variants are no longer ``out-of-distribution''). Surprisingly, we observe that for most models, the generalization gap is minimal (less than 1 percent). This indicates that the trained model, being exposed to a few language variants, became more robust.

\section{Conclusion}
In this study, we systematically evaluated the performance, generalization, and efficiency of various de-identification models across multiple scenarios, including across formats, language variants, and gendered names.

We observed that large LLMs are a degree of magnitude less efficient at de-identification compared to smaller LLM variants or BERT-based models. While large LLMs are generally more robust, more efficient models (e.g., BERT models) could be fine-tuned to better performance at very low cost, with only marginal improvements in performance increasing from 250 training notes to 1000 training notes. We found no disparity in model performance with respect to gendered names.

To improve the robustness of a popular and widely used publicly available BERT-based de-identification model \cite{obi_deid_bert_i2b2}, we developed BERT-MultiCulture-DEID, a fine-tuned BERT model that demonstrates improved generalization to identifiers from multiple language variants. 

\subsection{Future Work}
Our work has uncovered avenues for future work. The first and most direct step is to expand the set of language variants evaluated. Expanding the analysis to less-resourced languages is important to ensure equitable protection of privacy.  As newer models are released, their performance will need to be scrutinized in a similar way. We also need more publicly available benchmarking datasets to standardize and ensure transparency of future evaluations across studies.

We attempted to improve generalizability using an increased variety of data. Researchers can explore novel model architectures to improve the robustness of models without needing as much variety in the training data.

\section*{Limitations}
We acknowledge that our study has limitations. 
First, inference was performed under controlled hardware specifications (e.g., 2 GPUs). The findings (especially the specific measurements) are likely to change with other setups (e.g. faster CPUs or GPUs would lead to different results), though we would expect relative performance to stay the same.

Second, while the Faker library simplifies the replacement of PII in clinical notes, it uses fixed distributions for names, addresses, dates, and other identifiers. This may not fully capture real-world, multi-language diversity, potentially introducing biases during fine-tuning. Additionally, there is a disparity in the amount of variation available to different Faker locales (i.e., some languages only have a small pool of identifiers or no identifiers for specific identifier categories), which affects the overall distributions.

Another limitation stemming from the Faker library deals with gendered languages. Our analysis relies on Faker's pre-defined list which only deals with the male and female gender and does not meaningfully deal with the issues surrounding name-based gender identification. While we are aware of these issues, we believe that it is still vital to attempt to uncover any performance discrepancies which may negatively affect patient privacy. For this reason, we proceeded with the analysis.

Moreover, our evaluation of multi-cultural generalization does not cover the entire spectrum of language and cultural variations. Although there is a great deal of linguistic variety in the evaluated languages, we have not proven that our results necessarily generalize to all languages.

Another limitation of our approach is that clinical language and practices evolve over time, which could lead to degradation of our fine-tuned models without periodic updates or retraining. Similarly, applying these models to different datasets may result in differences in performance and ranking. 

Due to computational constraints (computational calculations are summarized in \ref{sec:compute}), we were also limited in the complexity and number of experiments that we could perform. Specifically, fine-tuning smaller LLM variant models (e.g., 8B parameters) was limited to a maximum of 2,000 training samples. We also could not test all models for all subexperiments due to the exorbitant cost associated with training larger LLMs. Thus, we limited our experiments to specific models that were most likely to be of use to the research community.

A further limitation is the presence of labeling errors in the MIMIC dataset. Some elements of PII are incorrectly categorized (e.g., identifiers such as name or organization were mislabeled). Despite labeling errors, the tokens still represent PII and are usable for de-identification. Past work manually evaluating the accuracy of the MIMIC data set's PII label found few errors, concluding that it would have no meaningful impact on their results \cite{aghakasiri2025not}. 

We believe that our work is generally low-risk as work serves to improve the robustness and efficiency of de-identification which in turn reduces the privacy risk of other works. We believe that the primary effect of our research is positive, though there are negative externalities with the execution of our work (e.g., the climate cost of running so many different models). Our work is limited in that we do not account for such externalities.

\section*{Acknowledgments}
Carrie Ye is supported by a CRAF (CIORA)-Arthritis Society Canada New Clinical investigator Award (award \#Cl-24-0013). Ross Mitchell is the Alberta Health Services Chair in Artificial Intelligence in Health and is supported by CIFAR, University Hospital Foundation, Amii, and the Canadian Foundation for Innovation. Mohamed Abdalla is supported by a CIFAR AI chair. Noopur Zambare is supported through an Amii grant.

\newpage
\bibliography{custom}

\newpage
\appendix

\section{Dataset Description}
\label{sec:dataset}
We sampled a total of 4,000 notes from MIMIC-III, which were divided into subsets for fine-tuning, validation, and evaluation. For fine-tuning, we experimented with training sets of varying sizes (250, 500, 1000, and 2000 notes) to study the effect of data scaling. The remaining notes were used for validation and final evaluation.  
\begin{table}[ht]
\begin{tabular}{l|cc} & \textbf{PCD} & \textbf{MIMIC-III} \\ \hline
Number of texts & 204 & 4000 \\ \hline
Note Length & & \\
\hspace{3mm}min  &  120& 66 \\
\hspace{3mm}mean & 670 &  1444\\
\hspace{3mm}max &  4975&   6680\\ \hline
\begin{tabular}[c]{@{}l@{}}Number of Sensitive\\ Words\end{tabular} & & \\
\hspace{3mm}min & 38 & 5 \\
\hspace{3mm}mean &  183 & 70 \\
\hspace{3mm}max & 981 &  362
\end{tabular}
\caption{Descriptive statistics of datasets used in this paper. \label{tab:data_stats}}
\end{table}

\section{PII Entity Distribution}

\begin{figure}[ht]
\includegraphics[width=\columnwidth]{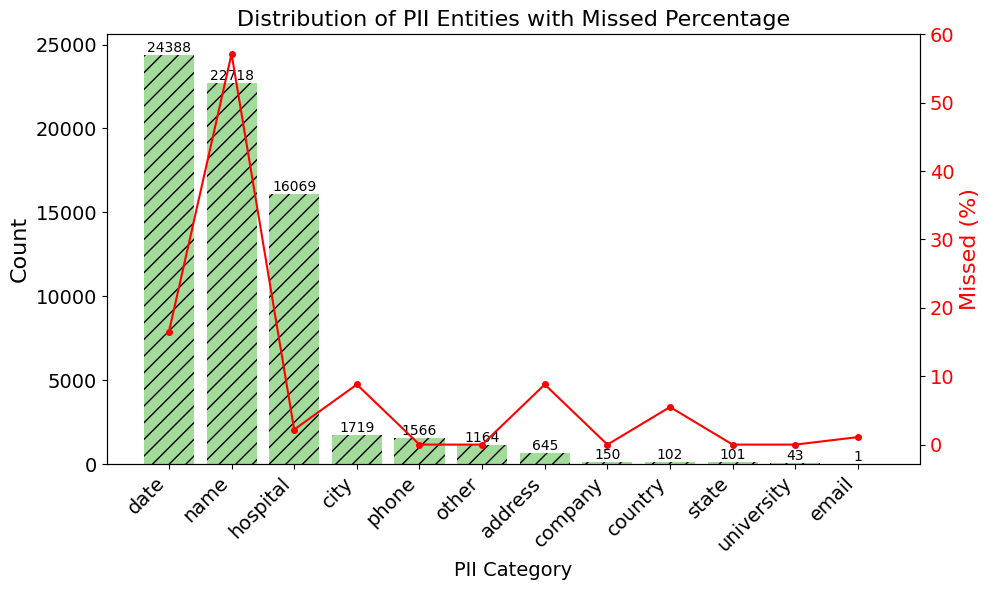}
\caption{Distribution of PII entities in the testing data. The left y-axis shows the total count of each PII type in the test set, and the right y-axis shows the percentage of missed identifiers out of all missed identifiers during de-identification by the best-performing model BERT.}
\label{fig:pii_distribution}
\end{figure}

Figure \ref{fig:pii_distribution} shows PII distribution in MIMIC-III test data.

\section{Fine Tuning Setup}
\label{sec:fine-tuning-setup}
We used Python-based libraries, transformers \citep{wolf-etal-2020-transformers} and PEFT \citep{peft} for fine-tuning. BERT, ClinicalBERT, and ModernBERT were fine-tuned for 10 epochs with a learning rate of $2 \times 10^{-5}$ and weight decay of 0.01. The batch size was 8 for BERT and ClinicalBERT, while it was 2 for ModernBERT. Due to their smaller context windows of 512, BERT and ClinicalBERT used a chunk size of 64, whereas ModernBERT, with a context window of 8192, used a chunk size of 1024. 

All Llama and Qwen models were LoRA fine-tuned with $\alpha = 16$ and $r = 8$. Llama-1, Qwen-1.5, and Llama-3 and Qwen-3 were fine-tuned for 5 epochs. Llama-8B and Qwen-7B were trained for 3 epochs. All these models used a batch size of 1, chunk size of 2048, and the same learning rate ($2 \times 10^{-5}$) and weight decay (0.01).

We also experimented with fine-tuning using different total numbers of samples, as summarized in Table~\ref{tab:performance-full}.

\begin{table}[t]
\centering
\tiny
\renewcommand{\arraystretch}{1.15}
\begin{tabular}{lcccc}
\hline
\textbf{Model} & \textbf{Fine Tuning Samples} & \textbf{Precision} & \textbf{Recall} \\
\hline
\multirow{4}{*}{BERT} 
                  & 250   & 0.995 & 0.996 \\
                  & 500     & 0.997 & 0.998      \\
                  & 1000    & 0.997 & 0.998       \\
                  & 2000  & 0.998 & 0.999   \\
\hline
\multirow{4}{*}{ClinicalBERT} 
                  & 250    & 0.989 & 0.995    \\
                  & 500   & 0.992 & 0.996   \\
                  & 1000   & 0.996 & 0.998     \\
                  & 2000    & 0.997 & 0.998    \\
\hline
\multirow{4}{*}{ModernBERT} 
                  & 250    & 0.990 & 0.992    \\
                  & 500   & 0.994 & 0.996    \\
                  & 1000  & 0.996 & 0.998  \\
                  & 2000  & 0.998 &   0.998   \\
\hline

\multirow{4}{*}{Qwen 1.5B} 
                  & 250 & 0.93 & 0.91     \\
                  & 500 & 0.94 & 0.93     \\
                  & 1000  & 0.96 & 0.94\\
                  & 2000  & 0.97 & 0.95   \\
\hline
\multirow{4}{*}{Qwen-3B} 
                  & 250 & 0.93 & 0.92     \\
                  & 500 & 0.95 & 0.94     \\
                  & 1000  & 0.96 & 0.95   \\
                  & 2000   & 0.97 & 0.96     \\
\hline
\multirow{4}{*}{Qwen-8B} 
                  & 250  & 0.92 & 0.92    \\
                  & 500  & 0.94 & 0.94      \\
                  & 1000   & 0.96 & 0.95    \\
                  & 2000 & 0.97 & 0.96  \\
\hline
\multirow{4}{*}{Llama-1B} 
                  & 250    & 0.96 & 0.94    \\
                  & 500      & 0.96 & 0.95  \\
                  & 1000  & 0.97 & 0.96    \\
                  & 2000    & 0.97 & 0.97   \\
\hline
\multirow{4}{*}{Llama-3B} 
                  & 250       & 0.95 & 0.95  \\
                  & 500     & 0.96 & 0.96    \\
                  & 1000    & 0.97 & 0.97    \\
                  & 2000    & 0.97 & 0.97       \\
\hline
\multirow{4}{*}{Llama-8B} 
                  & 250        & 0.95 & 0.96   \\
                  & 500     & 0.97 & 0.96      \\
                  & 1000   & 0.98 & 0.97       \\
                  & 2000    & 0.98 & 0.98  \\
          
\hline
\end{tabular}
\caption{Performance of various de-identification models fine-tuned on different numbers of MIMIC-III discharge summaries and evaluated on 1,000 MIMIC-III discharge summaries.}
\label{tab:performance-full}
\end{table}

\section{Multi-class Classification Results}
\label{sec:multi-class}
We also fine-tuned BERT variants and smaller models from the Llama and Qwen families as multiclass classifiers to de-identify multiple types of PII (for e.g., names, phone numbers, addresses, countries, city, hospital name, company name, and other identifiers). They were equally effective in performance as their binary classifier counterparts.

\label{sec:app-multi-class}

\begin{table}[t]
\tiny
\centering
\begin{tabular}{lll}
\hline
\textbf{Model}  & \textbf{P} & \textbf{R}  \\
\hline

 BERT               & 0.99 & 0.99 \\
 ModernBERT         & 0.99 & 0.99 \\
 ClinicalBERT       &   0.99 & 0.99 \\
 Qwen-1.5B           & 0.96& 0.94\\
 Qwen-3B               & 0.96 & 0.94 \\
 Qwen-7B                &0.96 &0.95 \\
 Llama-1B      &  0.97  & 0.95 \\
 Llama-3B      & 0.98 & 0.96 \\
 Llama-8B      &  0.98 & 0.97  \\
\hline
\end{tabular}
\caption{Performance of de-identification models fine-tuned as multiclass classifiers. \textbf{P: Precision, R: Recall}}
\label{tab:multi-claas}
\end{table}

\section{Prompts}
\label{sec:prompts}
\begin{figure*}[!htbp]
\centering
\includegraphics[width=0.70\textwidth]{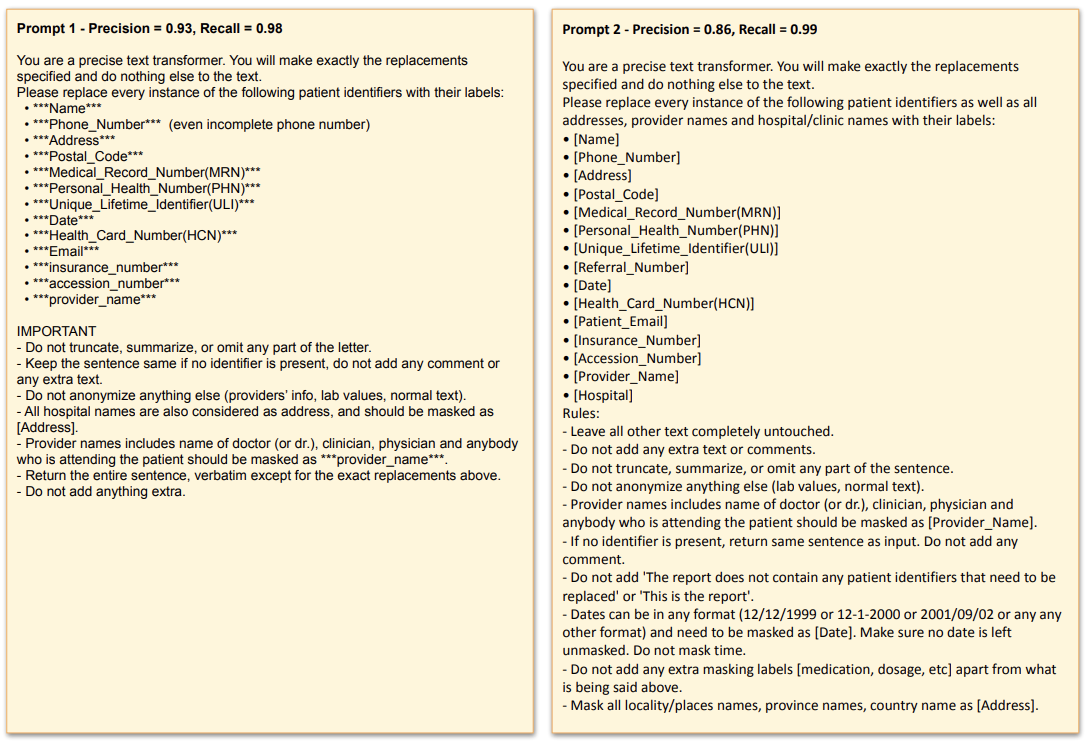}
\caption{Prompt-tuning Llama-70B with various prompts. \label{fig:prompt-tuning}}
\end{figure*}

\begin{figure*}[!htbp]
\centering
\includegraphics[width=0.70\textwidth]{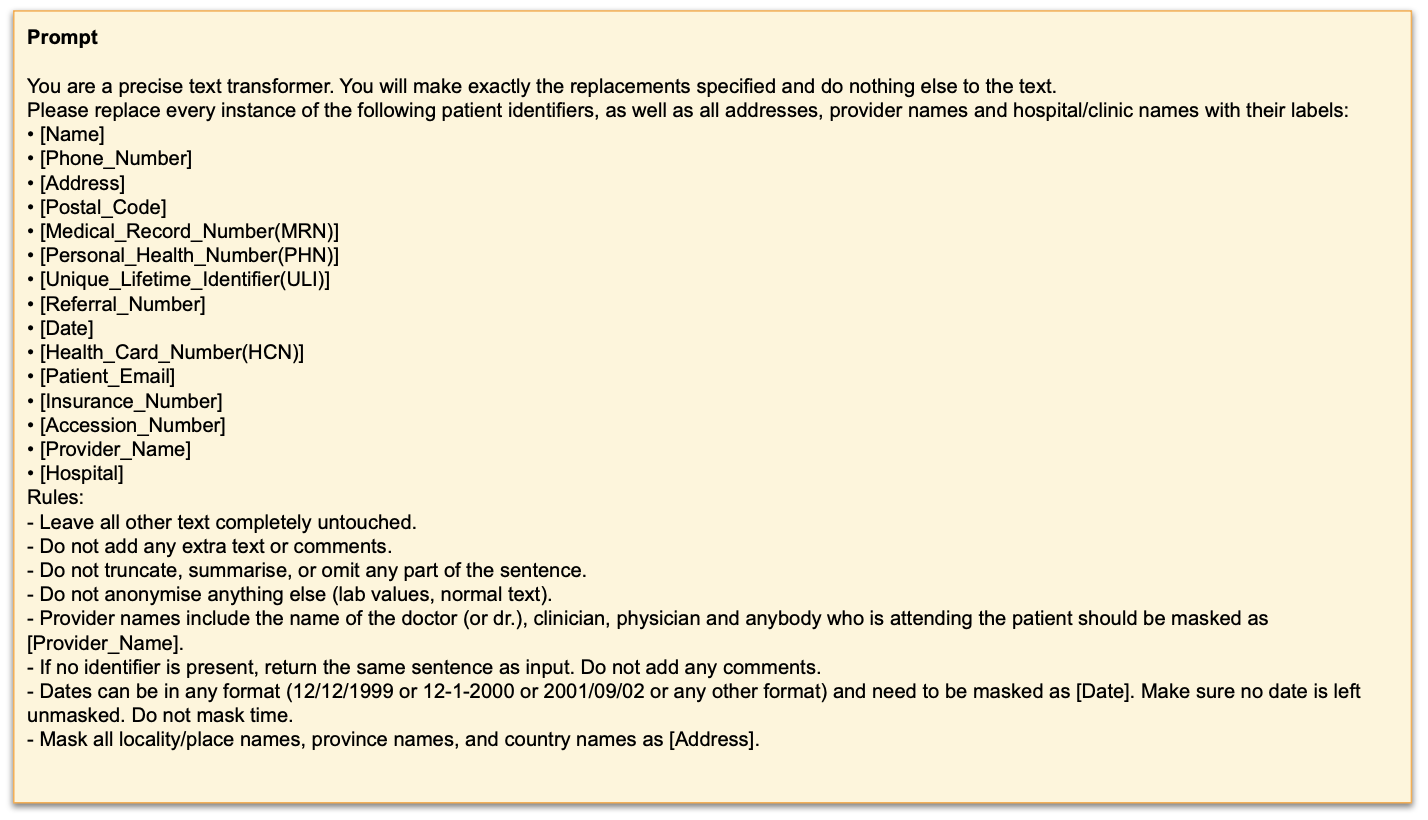}
\caption{Prompt used for de-identification using Llama-70B and Qwen-72B tuned on MIMIC dataset. \label{fig:prompt_mimic}}
\end{figure*}

\begin{figure*}[!htbp]
\centering
\includegraphics[width=0.70\textwidth]{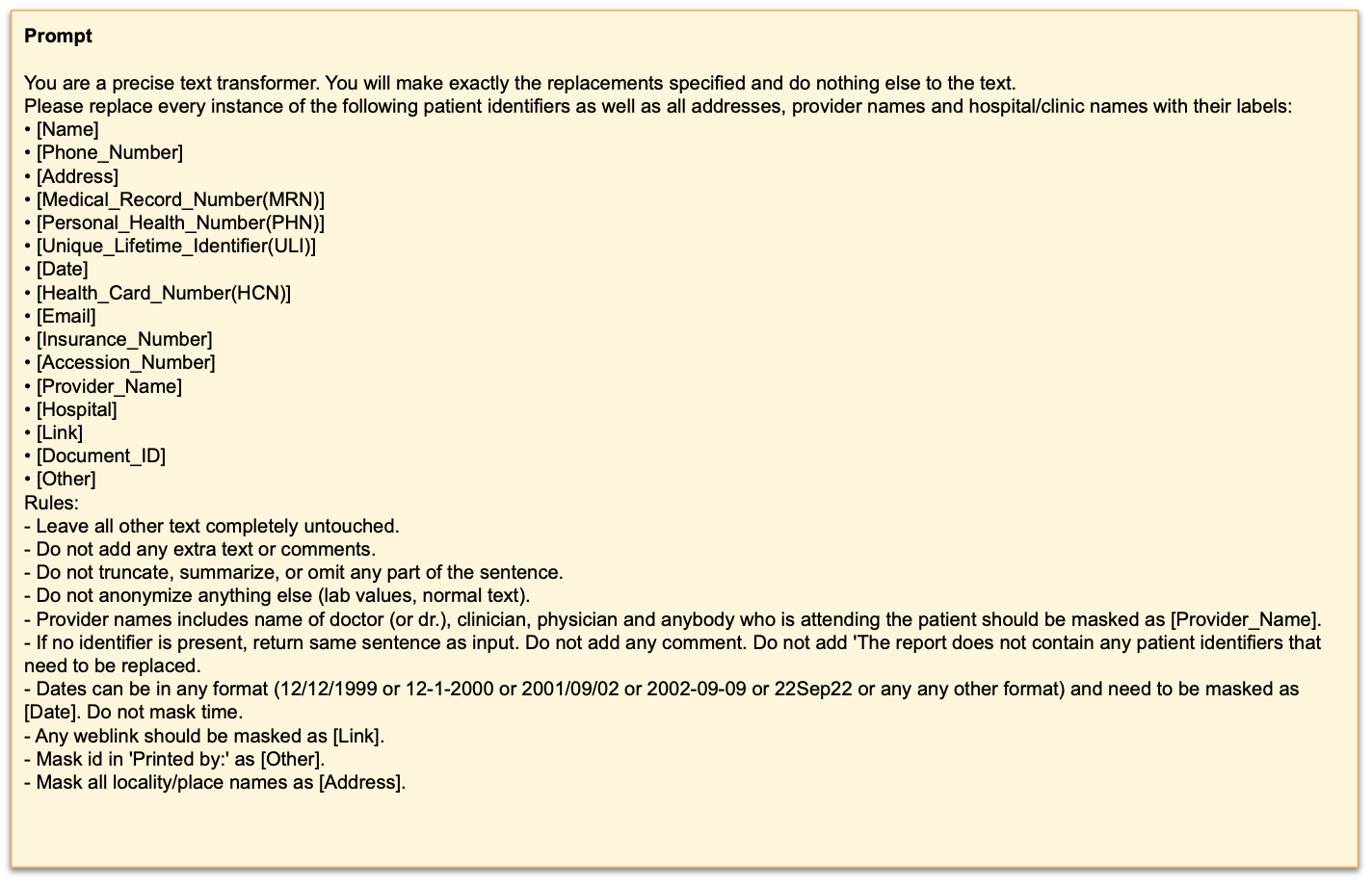}
\caption{Prompt used for de-identification using Llama-70B and Qwen-72B tuned on PCD. \label{fig:prompt_pcd}}
\end{figure*}

Our prompt-tuning process was conducted using 5 clinical notes. It began with an initial prompt synthesized from prior work. From there, we conducted an iterative refinement cycle: after running the model on 5 sample notes, we performed qualitative error analysis to identify errors (e.g., missing categories). Based on these observations, we made modifications to the prompt, focusing on improving recall while preserving precision. After each iteration (prompts and results for some iterations are given in Figure~\ref{fig:prompt-tuning}, we re-evaluated performance (on the same 5 notes); we repeated this process until further adjustments no longer yielded measurable recall improvements. The final prompt was then evaluated on the test set, and the result is reported in the paper.
Figure \ref{fig:prompt_mimic} presents the final prompt used for the MIMIC experiments with Llama-70B and Qwen-72B described in the paper. Figure \ref{fig:prompt_pcd} presents the prompt used for the experiments on the PCD dataset.

\begin{figure*}[!htbp]
\centering
\includegraphics[width=0.70\textwidth]{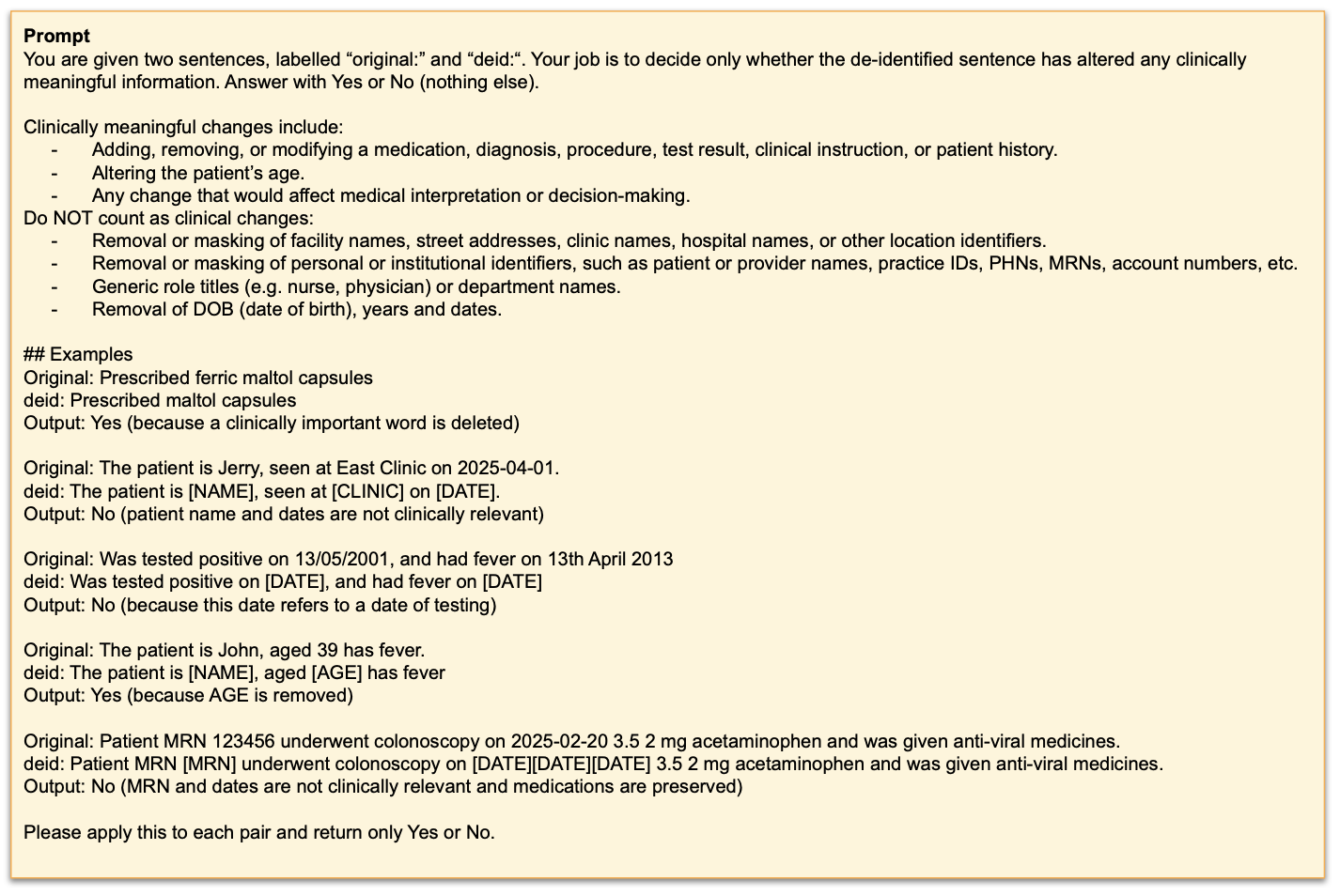}
\caption{Prompt used for CIRE with Llama-70B. \label{fig:CIRE}}
\end{figure*}

\section{Compute}
\label{sec:compute}
The experiments described in the paper were performed on NVIDIA A100-SXM4-80GB GPUs. All fine-tuning and inference experiments (except for Llama-70B and Qwen-72B) were conducted on a single GPU. Llama-70B and Qwen-72B required two GPUs. Fine-tuning all models took approximately 44 hours in total (for all sample sizes 250, 500, 1000 and 2000). Testing all fine-tuned models, including baselines, required around 31 hours. Prompt tuning and testing of Llama-70B and Qwen-72B took approximately 255 hours, while calculating CIRE required about 252 hours.

The parameter specifications of the models were as follows: BERT and ClinicalBERT contained 110M parameters, while ModernBERT contained 149M parameters. The Llama family consisted of models with 1B, 3B, 8B, and 70B parameters, and the Qwen family included models with 1.5B, 3B, 7B, and 72B parameters.

\section{Use of AI Assistants}
An AI assistant was used only for spelling, grammar, and phrasing.

\section{Full Results}
\label{app:full_results}
\begin{table*}[!htbp]
  \tiny
  \scriptsize
  \centering
  \begin{tabular}{lcc|cc|cc|cc|cc}
    \hline
    \textbf{Model} 
    & \multicolumn{2}{c}{\textbf{Mandarin}} 
    & \multicolumn{2}{c}{\textbf{Spanish}} 
    & \multicolumn{2}{c}{\textbf{Hindi}} 
    & \multicolumn{2}{c}{\textbf{French}}
    & \multicolumn{2}{c}{\textbf{Bengali}} \\
    \hline
    & P & R 
    & P & R 
    & P & R 
    & P & R 
    & P & R \\
    \hline
    BERT              & 0.99 & 0.80 & 0.99 & 0.77 & 0.99 & 0.80 & 0.99 & 0.83 & 0.99 & 0.75 \\
    ModernBERT     & 0.99 & 0.75 & 0.99 &0.83 & 0.99 & 0.92 & 0.99 & 0.90 &  0.99 & 0.81\\
    ClinicalBERT      & 0.99 & 0.87 & 0.99 & 0.84 & 0.99 & \textbf{0.97} & 0.99 & 0.87 & 0.99 & \textbf{0.97}\\
    \hline
    Qwen-1.5B & 0.97 & 0.80 & 0.96 & 0.86 & 0.96 & 0.91 & 0.96 & 0.89 & 0.96 & 0.80\\
    Qwen-3B & 0.97 & 0.65 & 0.96 & 0.82 & 0.97 & 0.90 & 0.97 & 0.89 & 0.97 & 0.89\\
    Qwen-7B & 0.98 & 0.71 &  0.97 & 0.85 & 0.97 & 0.89 & 0.97 & 0.89 & 0.97 & 0.76\\
    Llama-1b          & 0.98 & 0.83 & 0.97& 0.80& 0.97 & 0.90 &0.97 & 0.87 & 0.97 & 0.77 \\
    Llama-3b     & 0.98 & 0.71 &0.98 & 0.82 &0.98 & 0.88 & 0.98& 0.87  &0.97 & 0.76  \\
    Llama-8b     & 0.99 & 0.76 & 0.98 & 0.85 & 0.98 & 0.91 & 0.98 & 0.90 & 0.98 & 0.78     \\
    Llama70-prompting &  0.81 &  \textbf{0.96} &  0.79 &  \textbf{0.95} &  0.62 &  \textbf{0.97} & 0.74  &  \textbf{0.97} &  0.78 & 0.96  \\
    \hline
    pyDeid &  0.70 & 0.58 & 0.70 & 0.64 & 0.64 & 0.62 & 0.67 & 0.66 & 0.62 & 0.59\\
obi-deid-bert & 0.94 & 0.94 & 0.92 &0.84 &0.91 & 0.92 & 0.91 & 0.90 & 0.91& 0.93\\
Presidio & 0.44 & 0.94 & 0.64 & 0.90 & 0.60 & 0.87 & 0.64 & 0.92 & 0.58 & 0.85\\
    \hline
  \end{tabular}
  \caption{\textbf{Experiment 2.2.1:} Performance of de-identification models fine-tuned on 1000 MIMIC-III and tested on 500 samples from five languages. \textbf{P: Precision, R: Recall}}
  \label{tab:cross_lingual_test_appendix}
\end{table*}

\begin{table}[H]
  \scriptsize
  \centering
  \begin{tabular}{lcc|cc|cc}
    \hline
    \textbf{Model} & \multicolumn{2}{c}{\textbf{GB}} & \multicolumn{2}{c}{\textbf{AU}} & \multicolumn{2}{c}{\textbf{CA}} \\
    \hline
                   & P & R & P & R & P & R \\
    \hline
    BERT              &  0.99 & 0.97 & 0.99 & 0.98 & 0.99 & 0.99 \\
    ModernBERT        & 0.99 & 0.97 & 0.99 & 0.99 &  0.99 & 0.99 \\
    ClinicalBERT      & 0.99 & 0.96 & 0.99 & 0.99 & 0.99 & 0.99\\
    \hline
    Qwen-1.5B & 0.96 & 0.93 & 0.96 & 0.95 & 0.96 & 0.94\\
    Qwen-3B & 0.97 & 0.94 & 0.96 & 0.95 & 0.96 & 0.95\\
    Qwen-7B & 0.97 & 0.94 & 0.97 & 0.95 & 0.96 & 0.95\\
    Llama-1b          &  0.97 & 0.95 & 0.97 & 0.96 & 0.97 & 0.96 \\
    Llama-3b     &  0.97 & 0.96 & 0.97 & 0.97 & 0.97 & 0.96   \\
    Llama-8b     &  0.98 & 0.96 &  0.98 & 0.97 & 0.98 & 0.96 \\

    Llama-70B &  0.75 & 0.98 & 0.74 & 0.98 & 0.76 & 0.97 \\
    \hline
    pyDeid & 0.68 & 0.71 & 0.67 & 0.72 & 0.68 & 0.74\\
    obi-deid-bert & 0.91 & 0.91 & 0.91&0.94 & 0.91&0.93\\
    Presidio & 0.63 &0.88 & 0.62& 0.92 & 0.62 & 0.92\\
    \hline
  \end{tabular}
  \caption{\textbf{Experiment 2.2.2:} Performance of de-identification models fine-tuned on 1000 MIMIC-III and evaluated on 500 samples from above 3 English variants. \textbf{P: Precision, R: Recall}}
  \label{tab:cross_dialect_test}
\end{table}

\begin{table*}[!htbp]
  \tiny
  \centering
  \begin{tabular}{lcccccccccccccccc}
    \hline
    \textbf{Model} 
    & \multicolumn{2}{c}{\textbf{Mandarin}} 
    & \multicolumn{2}{c}{\textbf{Spanish}} 
    & \multicolumn{2}{c}{\textbf{Hindi}} 
    & \multicolumn{2}{c}{\textbf{French}}
    & \multicolumn{2}{c}{\textbf{Bengali}}
    & \multicolumn{2}{c}{\textbf{GB}}
\\
    \hline
&  $R_f$ &  $R_m$ &  $R_f$ &  $R_m$  &  $R_f$ &  $R_m$  &  $R_f$ &  $R_m$  &  $R_f$ &  $R_m$ & $R_f$ &  $R_m$  \\
    \hline
 BERT     & 0.92  &0.90 &0.88 &0.89 &0.58 &0.60 &\textbf{0.90} &\textbf{0.96} & 0.53& 0.52 & 1.0 & 1.0    \\
ModernBERT    & 0.85 & 0.86&0.94 & 0.93 &0.82 & 0.80& 0.96& 0.98& 0.76& 0.76& 0.99 & 0.99 \\
ClinicalBERT  & 0.85  & 0.85&0.96 & 0.95&0.93 & 0.95& 0.96&0.99 & 0.94& 0.97& 0.99  & 0.99\\
\hline
Qwen 1.5 & 0.92 & 0.91 & 0.81 & 0.83 & 0.78 & 0.82 & \textbf{0.81} & \textbf{0.87} & 0.70 & 0.72 & \textbf{0.90} & \textbf{0.95}\\
Qwen 3 & 0.81 & 0.83 & 0.80 & 0.83 & 0.80 & 0.83 & \textbf{0.83} & \textbf{0.88} &\textbf{0.71} & \textbf{0.76} &0.92 & 0.96\\
Qwen 7 & 0.87 & 0.84 & 0.82 & 0.84& 0.72 & 0.75 &\textbf{0.83} & \textbf{0.88} &0.62 & 0.64 & 0.92 & 0.95\\
Llama-1b     & 0.93 &0.90 & 0.78 & 0.81 & 0.72& 0.77& \textbf{0.82}& \textbf{0.88} & 0.68 &0.71 & \textbf{0.90} & \textbf{0.95}\\
Llama-3b       &  0.86 &0.85 & 0.81 & 0.83& 0.70 & 0.72 &\textbf{0.85} & \textbf{0.90}& 0.65 &0.70& \textbf{0.91}  &  \textbf{0.96}  \\
Llama-8b     & 0.93 &  0.90& 0.86 & 0.87&0.79 &0.82&0.87&0.91 &0.74&0.79 &0.92 &  0.96       \\
Llama-70B & 0.99 & 0.98 & 0.99 & 0.99 & 0.99& 0.99& 0.99 & 0.99 & 0.99 &0.99 & 0.99& 0.99\\
\hline
pyDeid &  \textbf{0.80} & \textbf{0.69 }& 0.78 & 0.75 & 0.63 & 0.63 & 0.72 & 0.70 & 0.63& 0.62 & \textbf{0.85} & \textbf{0.90}\\
obi-deid-bert & 0.98& 0.97& 0.91&0.91 &  0.91 & 0.94 &0.95 & 0.98 & 0.93& 0.94& 0.99&0.99\\
Presidio & 0.97 & 0.96 &0.88 &0.90 & 0.84 &0.84 & 0.93 & 0.95 & \textbf{0.72} & \textbf{0.80} & 0.97 & 0.97\\
    \hline
  \end{tabular}
  \caption{\textbf{Experiment 2.3:} Performance disparity across gendered names of de-identification models fine-tuned on 1000 MIMIC-III and tested on 500 samples having identifiers from six language variants.  Differences in recall greater than 0.05 between feminine and masculine names are highlighted. The Faker library does not provide an explicit pool for female and male names for Australian and Canadian English. \textbf{$R_f$: Female Recall, $R_m$: Male Recall}}
  \label{tab:cross_gender_test}
\end{table*}

\end{document}